# Improved Pothole Detection Using YOLOv7 and ESRGAN

Nirmal Kumar Rout, Gyanateet Dutta, Varun Sinha, Arghadeep Dey, Subhrangshu Mukherjee, Gopal Gupta

**Abstract**— Potholes are common road hazards that is causing damage to vehicles and posing a safety risk to drivers. The introduction of Convolutional Neural Networks (CNNs) is widely used in the industry for object detection based on Deep Learning methods and has achieved significant progress in hardware improvement and software implementations. In this paper, a unique better algorithm is proposed to warrant the use of low-resolution cameras or low-resolution images and video feed for automatic pothole detection using Super Resolution (SR) through Super Resolution Generative Adversarial Networks (SRGANs). Then we have proceeded to establish a baseline pothole detection performance on low quality and high quality dashcam images using a You Only Look Once (YOLO) network, namely the YOLOv7 network. We then have illustrated and examined the speed and accuracy gained above the benchmark after having upscaling implementation on the low quality images.

——————————◆——————————

## 1 INTRODUCTION

POTHOLES are a major issue on roads worldwide, causing damage to vehicles and posing a safety risk to drivers. Automated pothole detection systems can help to identify and repair potholes more efficiently, but the use of low-resolution cameras or low-quality video feed can be a challenge. In this paper, we propose a novel approach for improving the performance of pothole detection using low-resolution cameras or low-quality images and video feed. Our approach involves using an Enhanced Super Resolution Generative Adversarial Networks (ESRGAN) [1] to enhance the resolution of low-quality images and video feed, and then applying the You Only Look Once(YOLOv7) [2] object detection algorithm to detect potholes in the enhanced images. We compare the speed and accuracy of our approach to a baseline pothole detection system using YOLOv7 on high-quality images and show that it provides a significant improvement in both areas. We also demonstrate that our approach can be applied to a range of different road conditions and pothole types. One of the major advantages of our approach is its cost-effectiveness. ESRGAN can be used to improve the resolution of low-quality images and video feed from low-cost cameras, rather than requiring the use of high-resolution cameras with expensive sensors. This can greatly reduce the cost of implementing pothole detection systems, especially in resource-constrained settings. To validate the effectiveness of our approach, we conduct a series of experiments on a medium sized dataset of dash-cam images and video feed from a variety of international locations which indicate real life scenarios. Our results show that use of ESRGAN and YOLOv7 can significantly improve the performance of pothole detection systems and provide a reliable solution for detecting potholes in low-resolution images and video feed. This has the potential to greatly enhance the efficiency and effectiveness of pothole repair efforts and improve road safety for drivers worldwide.

————————————

- *Nirmal Kumar Rout is with School of Electronics Engineering, KIIT University, Bhubaneswar, India. Email: nkrout@kiit.ac.in.*
- *Gyanateet Dutta, Varun Sinha, Subhrangshu Mukherjee, Arghadeep Dey, Gopal Gupta are with School of Electronics Engineering, KIIT University, Bhubaneswar, India. E-mail: {1930198, 1930055, 1930053, 1930069, 1930020} @kiit.ac.in.*

## 2 RELATED WORKS

A number of approaches have been proposed in the literature for automated pothole detection. The earlier approaches [3] required 3-D equipment which can be very expensive and not suitable for use for all purposes. These techniques frequently use image data taken by digital cameras [4, 5] and depth cameras, thermal technology, and lasers. Recent approaches rely on machine learning algorithms and deep learning algorithms for image processing and detect potholes. Techniques based on Convolutional-neural-networks (CNN) are widely used for feature extraction of potholes from images because they can accurately model the non-linear patterns and perform automatic feature extraction and their robustness in separating unecessary noise and other image conditions in road images [6]. Even though, CNNs have been used in many approaches [7, 8, 9] they are ineffective in certain scenarios like while detecting objects which are smaller relative to the image. This can be solved by using high resolution images for detection but then the computational cost required for processing is too high, reason being CNNs are very memory consuming and they also require significantly high computation time. For addressing this issue, Chen et al. [10] suggested to using smaller input images or image patches from HR images for training the network. The first method is a two-phase system where a localization network (LCNN) is employed initially for locating frame segment of pothole in the image and then using a network for classification developed on part (PCNN) to calculate the classes. A recent study by Salcedo et al. [11] developed a road maintenance prioritization system for India using deep learning models such as UNet, which incorporates ResNet34 as the encoder, EfficientNet, and YOLOv5 on the Indian driving dataset(IDD). The study by Silva et al. [11], employed the YOLOv4 to detect damage on roads on a dataset of images taken from overhead view of an airborne drone. The study experimentally evaluated the accuracy and applicability of YOLOv4 in subject to recognizing highway road damages, and found an accuracy of 95%. The work proposed by Mohammad et al. [12] comprised of a system of using an edge platform using the AI kit(OAK-D) on frameworks such as the YOLOv1, YOLOv2, YOLOv3, YOLOv4, Tiny-YOLOv5, and SSD - mobilenet V2. In the work Anup et al. [13] proposed a 1D Convolutional Neural

Network to classify potholes using the data from gps information and accelerometer data collected using a smartphone device. Works by Amita et al. and Fan et al. [14, 15] solved the problem of annotation of data for training deep learning networks by proposing a pothole detection dataset based on stereo-vision and implemented stereo-vision analysis of road environments ahead of the vehicle by using a deep CNN. Another recent work by Salaudeen et al. [16] proposed an object detection solution based on YOLOv5 and using ESRGAN to upscale the images and have found better results compared to previous approaches. It still is very computationally expensive compared to how accurate it is and lacks flexibility in a variety of devices and would not be fit for every scenario. In our research, three different versions of YOLOv7 are utilized, viz. YOLOv7, YOLOv7x and YOLOv7 tiny. The standard version YOLOv7 features 75 convolutional layers and 36.9 M parameters. YOLOv7x is an upscaled version with 105 convolutional layers and 71.3 M parameters. These architectures are great for deploying in situations where there is edge based cloud computing. YOLOv7 tiny is a smaller version of the standard with 24 convolutional layers and 6.2 M parameters. This model is great for inferencing locally on edge devices which greatly benefit from its faster inference speed.

## 3 METHODOLOGY

In this section, we go through the several stages, including the collection and preprocessing of data, implementation of the ESRGAN and YOLOv7 algorithms and our procedure to evaluate results.

### 3.1 Dataset and Preprocessing

To ensure the accuracy and reliability of the variable outputs correlated with the prediction results using this deep learning model, we acquired sufficient non-linear image data required for training the network on. We assured that we had enough data to train the model, so that it performs well. The dataset used for this study was accessed free of charge from the internet from OpenCV's website. The dataset can be accessed via the link [17]. It consists of images from the dataset prepared by Electronic Department, Stellenbosch University. There are two folders in the dataset which are the images and labels. The images and labels folders have further sub folders namely Training, Testing and Validation where the labels folders have text files corresponding the coordinates of the bounding box of potholes in each image.The images folder has a total of 1784 images out of which 1265 images are for training, 401 images for validating and 118 images were reserved for testing providing us a ratio of 11:3:1. The images in the dataset are of different sizes and have different orientation of objects in them so they are resized to 1100 × 800 p. We also down sample the similar to 640 × 360 p to use for training without and use the following images for up scaling.We choose not to try to upscale the 800 p images because we ran into a bottleneck with the video memory of the graphics processing unit and we wanted to train the bigger network to a certain aimed number of epochs. Sample images with bounding box are shown in Fig.1. We use the PNW dataset [18] for comparative testing which is a video on YouTube that is taken on a highway in thePacific Northwest during the winter months as it simulates one of the real-world scenarios of detecting potholes on roads damaged by snowmelt and rain. The video footage was captured from a vehicle travelling at speeds between 45-90 km/h. The video frames are of size 1280×720.

### 3.2 Enhanced Super-Resolution Generative Adversarial Network (ESRGAN)

It is a deep learning model that is used for image super-resolution. It is based on the Generative Adversarial Network (GAN) [19] architecture and the Super Resolution Generative Adversarial Network (SRGAN) . We use the ESRGAN in our architecture because of its improvements over the other upscaling algorithms. The authors made adjustments to the discriminator and the perceptual loss component in the SRResNet [20] architecture with the aim of enhancing its performance. To improve the texture image quality of the output images, they made two changes to the ESRGAN's architecture. Firstly, they eliminated all layers for Batch Normalization (BN) and substituted them with a Residual-in-Residual Dense Block (RRDB) which combines multi-level residual networks and dense connections in one block, as shown in Fig.2. Additionally, the authors improved the the discriminator component of the architecture based on the Relativistic GAN [21]. The relativistic discriminator determines the relative realism realism of an image compared to a fake one, in comparison to the conventional discriminator in SRGAN, which accesses the probability of the input image being authentic , The authors, thus presented a suggestion for the generator network's discriminator loss and adversarial loss through the implementation of equations (1) and (2) as :

$$L_D^{R_a} = -E_{x_r}[log(D_{R_a}(x_r,x_f))] - E_{x_f}(log(1 - D_{R_a}(x_f,x_r))] \quad (1)$$

$$L_G^{R_a} = -E_{x_r}[log(1 - D_{R_a}(x_r,x_f))] - E_{x_f}[log(D_{R_a}(x_f,x_r))] \quad (2)$$

### 3.3 You Only Look Once (YOLOv7)

The YOLOv7 architecture builds up from the previous YOLO-R [22], YOLOv4 [23] and Scaled YOLOv4 [24] versions. But compared to older YOLOv4 and YOLOv3 [25] versions, which use Darknet and the YOLOv5 [26] uses PyTorch and CSPDarknet53 [27], the YOLOv7 has been trained entirely on the Microsoft COCO dataset [28], instead of using a pre-trained backbone from the ImageNet datatset. YOLOv7 is considered a cutting-edge solution in terms of efficiency and speed for tasks such as classification and object detection. It can process an entire image in one forward pass, making it a single-pass network . YOLOv7 introduces an Extended Efficient Layer Aggregation Network (E-ELAN) for computation in its backbone which is inspired from its predecessors to improve network efficiency. It considers the factors such as memory access cost, the I/O channel ratio, element wise operations, activations and the gradient path. It utilizes expand, shuffle, and merge techniques to increase the network's capacity to learn without disrupting and the original gradient path. YOLOv7 also supports compound model scaling which basically allows us to customize the architecture based on the application requirements to make able to fit in a wide range of computing devices by taking parameters like the resolution or size of the input image the model takes in, the width or the number of channels in the input that the model supports. The depth of the network meaning the number of layers in the network and its stage or the number of feature pyramids in the model.

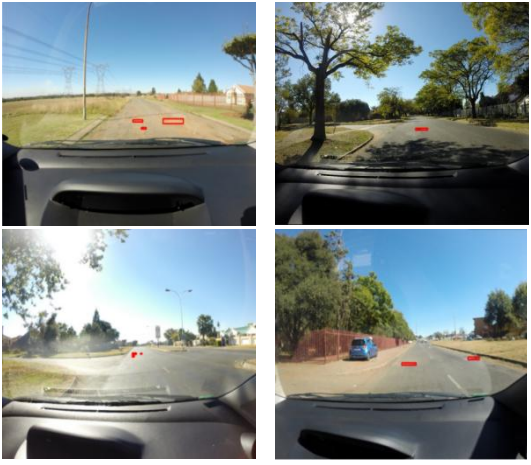

Fig 1. Sample dashcam pothole images with bounding boxes

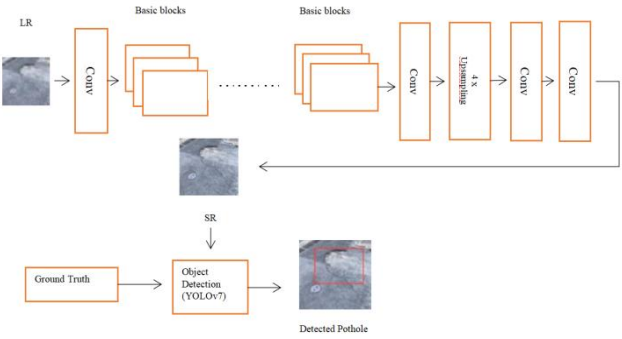

Fig 2. The suggested pipeline with the YOLOv7 object detector and the ESRGAN super-resolution upscaler.

### 3.4 Evaluation Metrics and Parameters

Many parameters were utilized while running the model. The attribute of object detection, which generally require to encircle interesting objects with bounding boxes with affirmed with careful assessment of the functioning of the module using three different metrics ( i.e, mAP, precision, and recall) shown in the Equations 3, 4 and 5.

$$mAP = \frac{1}{n}\sum_{k=1}^{k=n} AP_k \quad (3)$$

$$Precision = \frac{TP}{TP + FP} \quad (4)$$

$$Recall = \frac{TP}{TP + FN} \quad (5)$$

The definition of average precision for the number of the classes(n) and class K is represented by "$AP_k$". True Positives (TP) refer to the actual existence of an object in the image, while False Position (FP) indicate an incorrect inference made by the network to mark an item in which is not present. On the other hand False Negatives (FN) describe instances where an object exists in the image but goes undetected by the network, while True Negatives (TN) refer to the correct identification of an absence of an object in the image as illustrated in Table 1. The precision metric is calculated as the ratio of (TP) to positive classifications (TP + FP), while recall is calculated as the ratio of (TP) to actual positive instances (TP + FN). A high precision value implies that the detected objects are highly accurate, while a high recall value represents a low rate of missing important objects. The F1 score, which is the harmonic mean of precision and recall, is presented in equation 6. This study evaluates the model's results using these four metrics. When combined with a confusion matrix, the F1 score provides important information about the model's performance.

$$F1\ Score = \frac{2 * Precision * Recall}{Precision + Recall} \quad (6)$$

The assessment of the model's performance involves experimenting with different thresholds and recall values. The ($P_n$) and recall ($R_n$) are calculated by assuming N threshold being a combination of precision and recall values (n= 1,2,….,N). Average Precision (AP) is determined using Equation 7.

$$AP = \sum_{n=1}^{N}(R_n - R_{n-1})P_n \quad (7)$$

The Intersection over Union (IoU) metric assesses the level of overlap between the bounding boxes of a detected object (B) and its ground truth object (A) tby dividing the area of their intersection by the total area of both boxes. It serves as an indicator of the similarity between the two boxes.

$$IOU = \frac{|A \cap B|}{|A \cup B|} \quad (8)$$

TABLE 1
PREDICTION TRUTH TABLE

| Actual/ Prediction | Predicted as Positive | Predicted as Negative |
|---|---|---|
| **Positive** | True Positive(TP) | False Negative(FN) |
| **Negative** | False Positive(FP) | True Negative(TN) |

### 4 EXPERIMENTAL RESULTS AND DISCUSSION

A variety of YOLOv7 models were trained, including YOLOv7 tiny with multi resolution training, YOLOv7 normal with multi-resolution training and YOLOv7x with fixed resolution training using both super-resolution and low-resolution images. The new generated anchors are updated into the architecture using a YAML file. The pothole detector module was trained on NVIDIA GPUs to take advantage of the Pytorch framework which supports CUDA architecture and GPU hardware acceleration. We trained them on 1265 images of resolution 800p on a kaggle virtual machine with an Intel ® Xeon ® CPU @ 3GHz, 2× Tesla T4 GPUs with 15 GBs of VRAM each and 13 GBs of system memory 150 epochs with a batch size of 12. The hyper parameters for the training which included the learning rate was set to 0.01, the momentum for stochastic gradient descent was set to 0.934, the weight decay to 0.0005 and we considered 3 warm epochs with the warm up momentum as 0.8 and the warm up bias learning rate as 0.1. At the start, the loss for all the architectures is high but after 20 epochs becomes less than 0.5 and converges towards 0 with further increase in epochs as shown in Fig. 3. The rate of the accuracy of the models are shown in Table 2. In this study, we found out that the YOLOv7x performed the best out of all while trained on low resolution images and also performed best with super resolution images also beating the performance on the low resolution images plus detecting more potholes. The other models performed average but they had better results on super resolution images compared to the low resolution images. We considered only the multi resolution models of the other architectures other than YOLOv7x because they represent best performance scenario. It is found that there is a trade off between using different architectures. However, the YOLOv7x and YOLOv7 models are very accurate, due to their larger number of parameters (around 71.3M and 36.9M respectively, still an improvement over previous state of the art architectures which have even more larger parameters) and their more demanding computational requirements ended up being slower for inference in real time compared to the YOLOv7 tiny which has around 6.2 M parameters and requires only 5.8 FLOPS compared to 104 FLOPS and 190 FLOPS on the YOLOv7 and YOLOv7x.

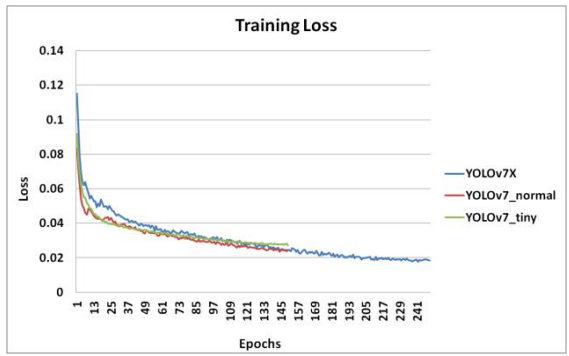

Fig 3. Loss of YOLOv7x, YOLOv7 multi res and YOLOv7 tiny multi res in the training process

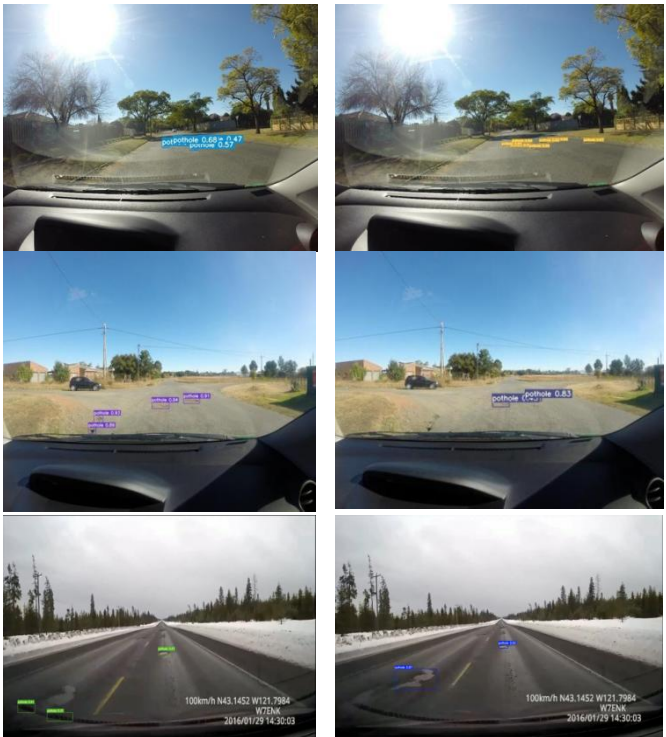

A) LR images    B) SR images

Fig 4. Detection result on images from two datasets.

TABLE 2
RESULTS OF MODELS ON COMBINED DATASET

| Arch | Img Res | mAP @ 0.5 | mAP@ 0.5:0.95 | Prec | Rec | Inf(in s) | F1 |
|---|---|---|---|---|---|---|---|
| YOLOv7x+ ESRGAN | LR | 0.73 | 0.39 | 0.80 | 0.68 | 0.053 | 0.735 |
|  | SR | 0.85 | 0.48 | 1.00 | 0.89 | 0.069 | 0.942 |
| YOLOv7 multi + ESRGAN | LR | 0.69 | 0.36 | 0.79 | 0.67 | 0.012 | 0.725 |
|  | SR | 0.81 | 0.45 | 0.98 | 0.85 | 0.018 | 0.910 |
| YOLOv7 tiny Multi + ESRGAN | LR | 0.73 | 0.36 | 0.77 | 0.67 | 0.006 | 0.717 |
|  | SR | 0.78 | 0.42 | 0.95 | 0.82 | 0.009 | 0.880 |

Due to the higher accuracy of the YOLOv7x model we further performed detection results on it. The YOLO models trained with super resolution images not only increased the accuracy and precision but also increased the number of potholes it could detect, including potholes in further areas. We show sample detection images in Fig. 4, where we can see that the model trained on super resolution images performs better. In our study, we employed the IoU threshold of 0.5 to 0.9 to calculate the recall. To access the efficacy of our models and compare their performance against currently leading results, precision and recall are commonly used as evualation metrics for classification. We already have discussed the metrics previously. We also consider and compare the inference time as a metric because when it comes to real time detection, a faster inference would give us better performance. We assessed the performance of the proposed object detectors and discovered that they were effective in identifying pothole instances and achieved comparable results. When it came to the inference time, we saw faster results, about two times when compared to the previous models on the same hardware specification. A selection of frames from the PNW dataset were analyzed and the results are displayed in Table 3. These frames effectively recreate the scenario of a vehicle in rapid motion and present realistic road damage obstacles. Here we compare our models with Dhiman and Klette's [15] top-performing method (LM1) model and Salaudeen and Çelebi's [16] top performing method (ESRGAN + EfficientDet) on extracted frames from the PNW dataset. The YOLOv7 tiny multi resolution model has an overall precision of 0.947 (94.7 %) and a recall of 0.826 (82.6 %) with an inference speed of 0.0093 seconds. In the Table 3, a comparison of our results with those of previous studies revealed that our methods achieved higher precision and recall scores than recent state of art methods. Our proposed techniques for identifying potholes outperform current methods by demonstrating a higher capability in detecting smaller potholes within the frame or from a distance, leading to a general enhancement in detection accuracy. Additionally, this method does not require expensive sensors such as Light Detection and Ranging (LIDAR) sensors or HD (High Definition) cameras to obtain accurate results.

TABLE 3
COMPARISONS WITH OTHER STATE-OF-THE-ART METHODS

| Authors | Method | Mean Prec | Recall | Inf(in s) |
|---|---|---|---|---|
| **Dhiman and Klette** | LM1 | 0.886 | 0.850 | 0.0945 |
| **Salaudeen and Çelebi** | EffecientDet + ESRGAN | 1.00 | 0.63 | 1.0317 |
| **Salaudeen and Çelebi** | YOLOv5 + ESRGAN | 0.925 | 0.861 | 0.0281 |
| **Our Proposed Model** | YOLOv7x + ESRGAN | 1.00 | 0.893 | 0.0696 |
| **Our Proposed Model** | YOLOv7 multi +ESRGAN | 0.984 | 0.867 | 0.0189 |
| **Our Proposed Model** | YOLOv7 tiny multi +ESRGAN | 0.947 | 0.826 | 0.0093 |

## 4.1 Limitations

The findings of our experiments revealed that the techniques employed by us outperformed previous methods in terms of precision and recall. This implies that the methods have a higher accuracy of detecting potholes, including small ones. However, the recall value was comparatively lower, to the accuracy of our own models, which could be due to the misidentification in the training and testing datasets. This means that while the methods can detect many potholes, they miss a few instances of them.

It was noted that the dataset had a significantly small but still notable number of mislabels, where cracks were identified as potholes or non-damaged road surfaces. This may be due to factors such as low image resolution during annotation, the multi-class nature of the dataset, and visual occlusions making the potholes difficult to detect. The YOLOv7 tiny model suffered more from this issue since it has the least number of layers compared to the bigger models we trained. Furthermore, computing resources also influenced the detection performance. Resizing the input images was necessary to train on high-resolution images, but the limited number of images and training time limited our results. The experiments demonstrate that improved evaluation metrics could be achieved with better-labelled dataset that more accurately identifies instances of potholes.

## 5 CONCLUSION

In this research, a method relying on enhancing the resolution of images was proposed as a solution to the difficulties in detecting potholes, a significant factor in road accidents, vehicle damage and tire degradation. The GAN-based ESRGAN network was used to enhance low-resolution and three object detectors, YOLOv7x, YOLOv7 multi resolution and YOLOv7 tiny multi resolution, were employed to identify potholes from improved images. The technique presented in this research demonstrated an improvement over existing methods in detecting small and distant potholes, as well as objects in challenging conditions. The object detection techniques produced better results with YOLOv7 tiny-multi being significantly faster during training and inference. This study also highlights the potential for using super-resolution images in other areas of research and future plans include end-to-end training and developing lightweight networks to improve efficiency. We plan to work on a pothole detection framework which uses an ESRGAN with a future release of YOLO on a system on chip and a camera which houses a neural accelerator that can run the ESRGAN in real time to upscale low resolution image frames in and then pass it through the YOLO network which would result in outstanding performance in all scenarios.